\documentclass[conference]{IEEEtran}
\IEEEoverridecommandlockouts
\usepackage{booktabs}
\usepackage{multirow}
\usepackage{makecell}

\usepackage{amsfonts}
\usepackage{amsmath}
\usepackage{subfigure}
\usepackage{diagbox}
\usepackage{tikz}
\usepackage[ruled,vlined,linesnumbered]{algorithm2e}

\usepackage{cite}
\usepackage{amsmath,amssymb,amsfonts}
\usepackage{algorithmic}
\usepackage{graphicx}
\usepackage{textcomp}
\usepackage{xcolor}
\def\BibTeX{{\rm B\kern-.05em{\sc i\kern-.025em b}\kern-.08em
    T\kern-.1667em\lower.7ex\hbox{E}\kern-.125emX}}
\begin{document}

\title{Innate-Values-driven Reinforcement Learning \\ based Cooperative Multi-Agent Cognitive Modeling
% {\footnotesize \textsuperscript{*}Note: Sub-titles are not captured for https://ieeexplore.ieee.org  and
% should not be used}
% \thanks{Identify applicable funding agency here. If none, delete this.}
}

\author{\IEEEauthorblockN{Qin Yang}
\IEEEauthorblockA{\textit{Intelligent Social Systems and Swarm Robotics Lab (IS$^3$R)} \\
\textit{Computer Science and Information Systems Department}\\
Bradley University, Peoria, USA \\
email: is3rlab@gmail.com}
% \and
% \IEEEauthorblockN{2\textsuperscript{nd} Given Name Surname}
% \IEEEauthorblockA{\textit{dept. name of organization (of Aff.)} \\
% \textit{name of organization (of Aff.)}\\
% City, Country \\
% email address or ORCID}
% \and
% \IEEEauthorblockN{3\textsuperscript{rd} Given Name Surname}
% \IEEEauthorblockA{\textit{dept. name of organization (of Aff.)} \\
% \textit{name of organization (of Aff.)}\\
% City, Country \\
% email address or ORCID}
% \and
% \IEEEauthorblockN{4\textsuperscript{th} Given Name Surname}
% \IEEEauthorblockA{\textit{dept. name of organization (of Aff.)} \\
% \textit{name of organization (of Aff.)}\\
% City, Country \\
% email address or ORCID}
% \and
% \IEEEauthorblockN{5\textsuperscript{th} Given Name Surname}
% \IEEEauthorblockA{\textit{dept. name of organization (of Aff.)} \\
% \textit{name of organization (of Aff.)}\\
% City, Country \\
% email address or ORCID}
% \and
% \IEEEauthorblockN{6\textsuperscript{th} Given Name Surname}
% \IEEEauthorblockA{\textit{dept. name of organization (of Aff.)} \\
% \textit{name of organization (of Aff.)}\\
% City, Country \\
% email address or ORCID}
}

\maketitle

\begin{abstract}
In multi-agent systems (MAS), the dynamic interaction among multiple decision-makers is driven by their innate values, affecting the environment's state, and can cause specific behavioral patterns to emerge. On the other hand, innate values in cognitive modeling reflect individual interests and preferences for specific tasks and drive them to develop diverse skills and plans, satisfying their various needs and achieving common goals in cooperation. Therefore, building the awareness of AI agents to balance the group utilities and system costs and meet group members' needs in their cooperation is a crucial problem for individuals learning to support their community and even integrate into human society in the long term. However, the current MAS reinforcement learning domain lacks a general intrinsic model to describe agents' dynamic motivation for decision-making and learning from an individual needs perspective in their cooperation. To address the gap, this paper proposes a general MAS innate-values reinforcement learning (IVRL) architecture from the individual preferences angle. We tested the Multi-Agent IVRL Actor-Critic Model in different StarCraft Multi-Agent Challenge (SMAC) settings, which demonstrated its potential to organize the group's behaviours to achieve better performance.

% Innate values describe agents' intrinsic motivations, which reflect their inherent interests and preferences to pursue goals and drive them to develop diverse skills satisfying their various needs. The essence of reinforcement learning (RL) is learning from interaction based on reward-driven (such as utilities) behaviors, much like natural agents. It is an excellent model to describe the innate-values-driven (IV) behaviors of AI agents. Especially in multi-agent systems (MAS), building the awareness of AI agents to balance the group utilities and system costs and satisfy group members' needs in their cooperation is a crucial problem for individuals learning to support their community and integrate human society in the long term. This paper proposes a hierarchical compound intrinsic value reinforcement learning model -- innate-values-driven reinforcement learning termed IVRL to describe the complex behaviors of multi-agent interaction in their cooperation. We implement the IVRL architecture in the StarCraft Multi-Agent Challenge (SMAC) environment and compare the cooperative performance within three characteristics of innate value agents (Coward, Neutral, and Reckless) through three benchmark multi-agent RL algorithms: QMIX, IQL, and QTRAN. The results demonstrate that by organizing individual various needs rationally, the group can achieve better performance with lower costs effectively.

\end{abstract}

% Uncomment the following to link to your code, datasets, an extended version or similar.
%
% \begin{links}
%     \link{Code}{https://aaai.org/example/code}
%     \link{Datasets}{https://aaai.org/example/datasets}
%     \link{Extended version}{https://aaai.org/example/extended-version}
% \end{links}

\section{Introduction}
The study of motivation has been ongoing for years in natural intelligence systems, which have developed a broad spectrum of different theories \cite{heckhausen2018motivation}. From the individual perspective, an agent's intrinsic motivation falls in the category of cognitive motivation theories, including theories of the mind that tend to abstract ideas from the biological system of the behaving organism \cite{merrick2013novelty}. It is concerned explicitly with the activities of creatures that reflect the pursuit of a particular goal and form a meaningful unit of behavior in this function \cite{heckhausen2018motivation}. Furthermore, intrinsic motivations describe incentives relating to an activity itself, and these incentives residing in pursuing an activity are intrinsic. They derive from an activity driven primarily by interest or activity-specific incentives, depending on whether the object of an activity or its performance provides the main incentive \cite{schiefele1996motivation}. 

In cognitive science, a significant portion of human activities involves acquiring and using skills and strategies driven by their intrinsic motivations and needs, which is naturally a primary focus of research and widely studied in cognitive skill acquisition \cite{vanlehn1996cognitive}.
Particularly, skills vary in complexity and degree of cognitive involvement, ranging from simple motor movements and routine everyday activities to high-level intellectual strategies \cite{sun2001implicit}. Specifically, in artificial intelligence (AI) methods, a strategy describes the general plan of an AI agent (like robot) achieving short-term or long-term goals under conditions of uncertainty, which involves setting sub-goals and priorities, determining action sequences to fulfill the tasks, and mobilizing resources to execute the actions \cite{freedman2015strategy}.

\begin{figure}[t]
\centering
\includegraphics[width=1\columnwidth]{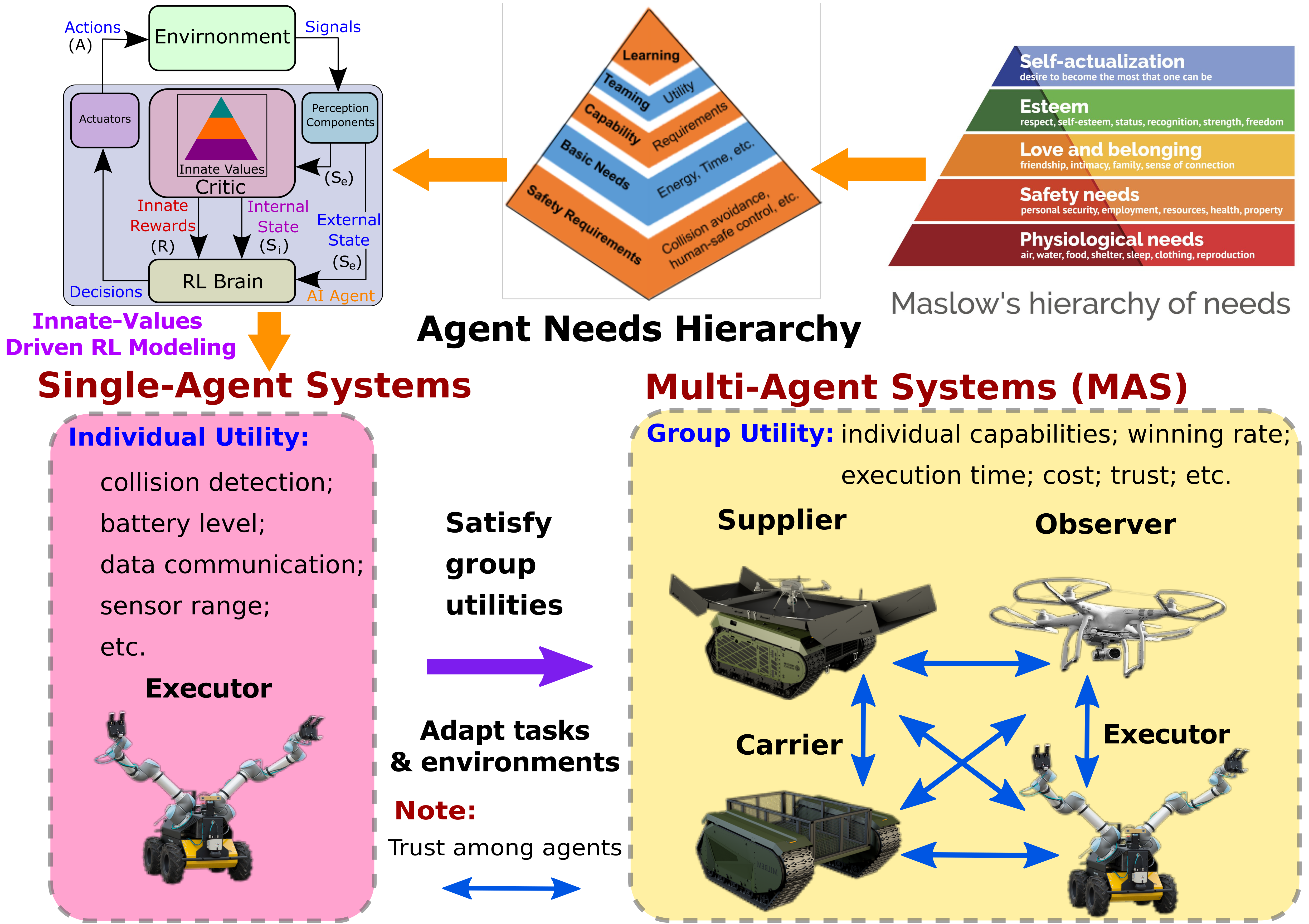}
\caption{The model derivation from Maslow's Hierarchy of Needs to innate-values-driven RL in the AI agent setting and relationships between the single-agent and multi-agent (like mobile robots such as UGV and UAV) in their cooperation.}
\label{fig:overview}
\end{figure}

% they are computational systems in which multiple autonomous agents work together to perform tasks to satisfy goals. These systems might be homogenous or heterogeneous and involve agents having a common goal or distinct or contradictory goals, considering varying degrees of communication during their interaction.
From the MAS perspective, the main issues include how to develop coordination strategies (that enable groups of agents to solve problems together effectively), negotiation mechanisms (that serve to bring a set of agents together), conflict detection and resolution strategies, and other mechanisms whereby agents can contribute to overall system effectiveness while still assuming autonomy.
To tackle those problems, cognitive modeling provides a more realistic basis for understanding multi-agent interaction by embodying realistic constraints (like agents' innate-values and needs), capabilities, and tendencies of individual agents in interacting with their environments, including physical and social environments \cite{sun2001implicit}. Fig. \ref{fig:overview} illustrates the derivation of models from the individual AI agent needs hierarchy to swarm intelligence and describes the relationships between the single-agent and cooperative multi-agent systems, such as multi-robot systems (MRS).

Specifically, researchers propose various abstract computational structures to form the fundamental units of cognition and motivations, such as states, goals, actions, and strategies. For intrinsic motivation modeling, the approaches can be generally classified into three categories: prediction-based \cite{schmidhuber1991curious}, novelty-based \cite{marsland2000real}, and competence-based \cite{schembri2007evolution}. Moreover, most MAS implementations aim to optimize the system's policies with respect to individual needs and intrinsic values, even though many real-world problems are inherently multi-objective \cite{ruadulescu2020multi}. Thus, many conflicts and complex trade-offs in the MAS need to be managed, and compromises among agents should be based on the utility mapping the innate values of a compromise solution -- how to measure and what to optimize \cite{zintgraf2015quality}. However, in the MAS setting, the situations will become much more complex when we consider individual utility reflects its own needs and preferences \cite{yang2020hierarchical}. 
For example, although we assume each group member receives the same team rewards in fully cooperative MAS, the benefits received by an individual agent are usually significant differences according to its contributions and innate values in real-world scenarios or general multi-agent settings. 
% However, in real scenarios, the rewards are generated by agents' innate value systems, which differ vastly from individuals based on their needs and requirements. 
In other words, considering the AI agent as a self-organizing system (like robots), developing its awareness through balancing internal and external utilities based on its needs in different tasks is a crucial problem for individuals learning to support others and integrate community with safety and harmony in the long term. 
\begin{figure}[t]
\centering
\includegraphics[width=\columnwidth]{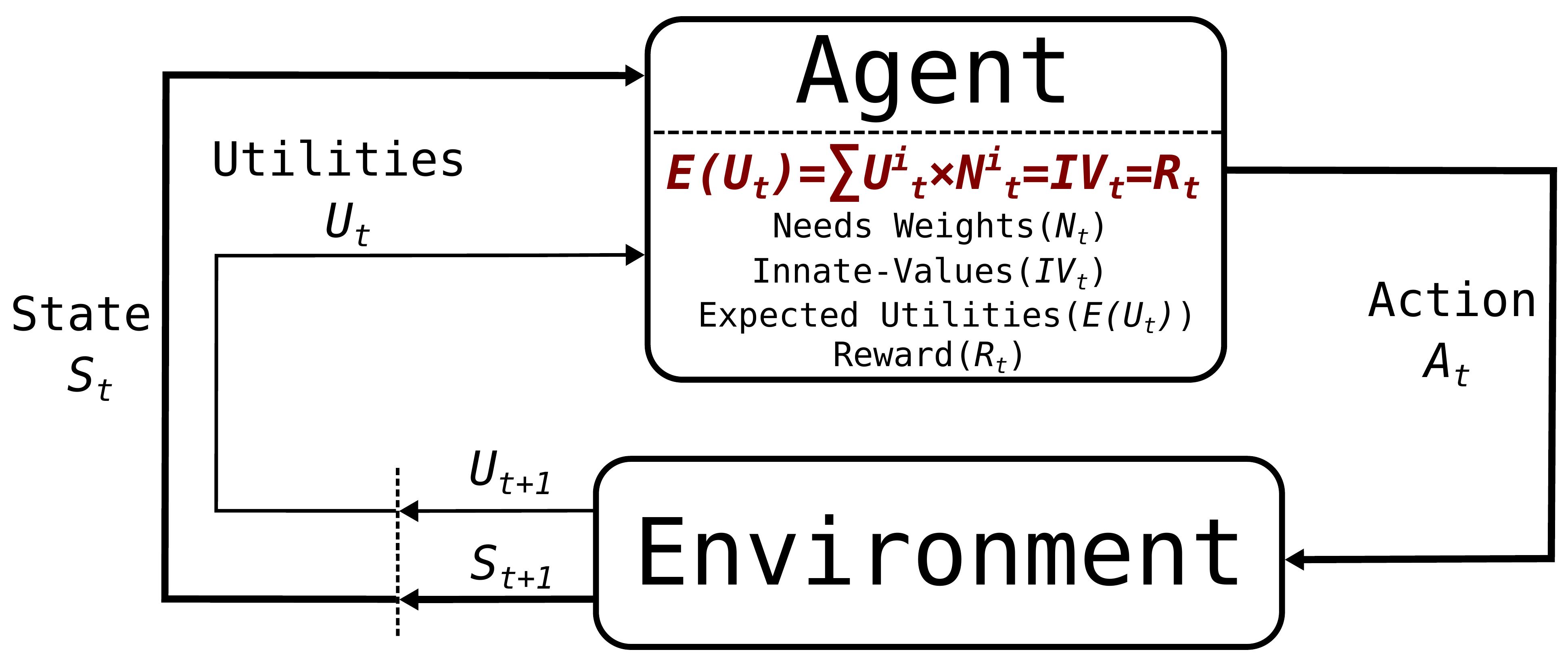}
\caption{The illustration of the innate-values-driven reinforcement learning (IVRL) single-agent model.}
\label{fig:single_ivrl_single_agent}
\end{figure}
\begin{figure}
\begin{center}
% \begin{tikzpicture}[main/.style = {draw, circle}] 
\begin{tikzpicture}[node distance={9.3mm}]
% \node[main] (1) {$s_1$}; 
\node (1) {$s_1$}; 
\node (2) [below right of=1] {$w_1$}; 
\node (3) [above right of=2] {$a_1$}; 
\node (4) [below right of=3] {$r_1$}; 
\node (5) [above right of=4] {$s_2$}; 
\node (6) [below right of=5] {$w_2$}; 
\node (7) [above right of=6] {$a_2$}; 
\node (8) [below right of=7] {$r_2$}; 
\node (9) [above right of=8] {$s_3$}; 
\node (10) [below right of=9] {$w_3$}; 
\node (11) [above right of=10] {$a_3$}; 
\node (12) [below right of=11] {$r_3$}; 
\node (13) [above right of=12] {$\cdots$}; 

\draw[->] (1) -- (2);
\draw[->] (1) -- (3);
\draw[->] (2) -- (4);
\draw[->] (3) -- (4);
\draw[->] (3) -- (5);
\draw[->] (5) -- (6);
\draw[->] (5) -- (7);
\draw[->] (6) -- (8);
\draw[->] (7) -- (8);
\draw[->] (7) -- (9);
\draw[->] (9) -- (10);
\draw[->] (9) -- (11);
\draw[->] (10) -- (12);
\draw[->] (11) -- (12);
\draw[->] (11) -- (13);
\end{tikzpicture}
\end{center}
\caption{Illustration of the trajectory of state $S$, needs weight $W$, action $A$, and reward $R$ in the IVRL model.}
\label{fig:iv_tr}
\end{figure}
\begin{figure}[t]
\centering
\includegraphics[width=\columnwidth]{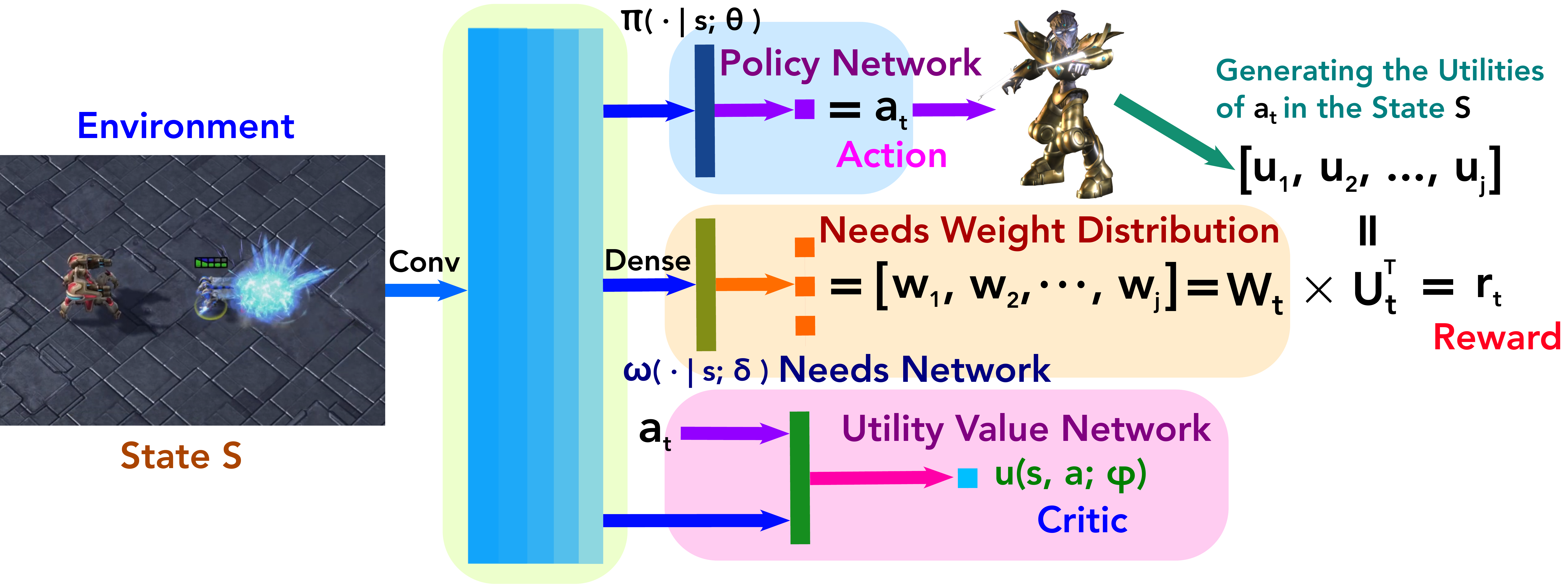}
\caption{The illustration of the innate-values-driven reinforcement learning (IVRL) with actor-critic model in an agent (Zealot) of the StarCraft II.}
\label{fig:single_ivrl_a2c}
\end{figure}

To address those gaps, we proposes a general MAS innate-values reinforcement learning (IVRL) architecture from the single agent's preferences angle to mimic the complex behaviors in MAS cooperation. We tested the Multi-Agent IVRL Actor-Critic Model in different StarCraft Multi-Agent Challenge (SMAC) settings, which demonstrated the potential to organize the group's behaviours to achieve better performance.

\section{Background and Preliminaries}

This section briefly reviews the expected utility theory, innate-values-driven (IV) behaviours modeling, and multi-Agent policy gradient learning. When describing a specific method, we use the notations and relative definitions from the corresponding papers.

\subsection{Expected Utility based IV Behaviours Modeling}

Unlike the traditional RL, innate-values-driven RL (IVRL) is based on combined motivations models and expected utility theory to mimic its complex behaviors in the evolution through decision-making and learning \cite{yang2024rationality}.
In the IVRL model, the reward is equal to the sum of each category's needs weight $n_k$ times its current utility $u_k$, which can also be regarded as the {\it expected utility} \cite{fishburn1979utility,fishburn1988nonlinear} generated by the agent's utility values $u(x_k)$ and the corresponding probability $p_k$ (Eq. \eqref{reward}).
\begin{equation}
R_t = \sum_{i=1}^{k} u_k \times n_k = \mathbb{E} \left[ U(p) \right] = \sum_{i=1}^k u(x_k) p_k
\label{reward}
\end{equation}

\begin{figure*}[t]
\centering
\subfigure[IVDAC-sum structure]{
\label{fig:ivadc_sum}
\includegraphics[width=0.8\columnwidth]{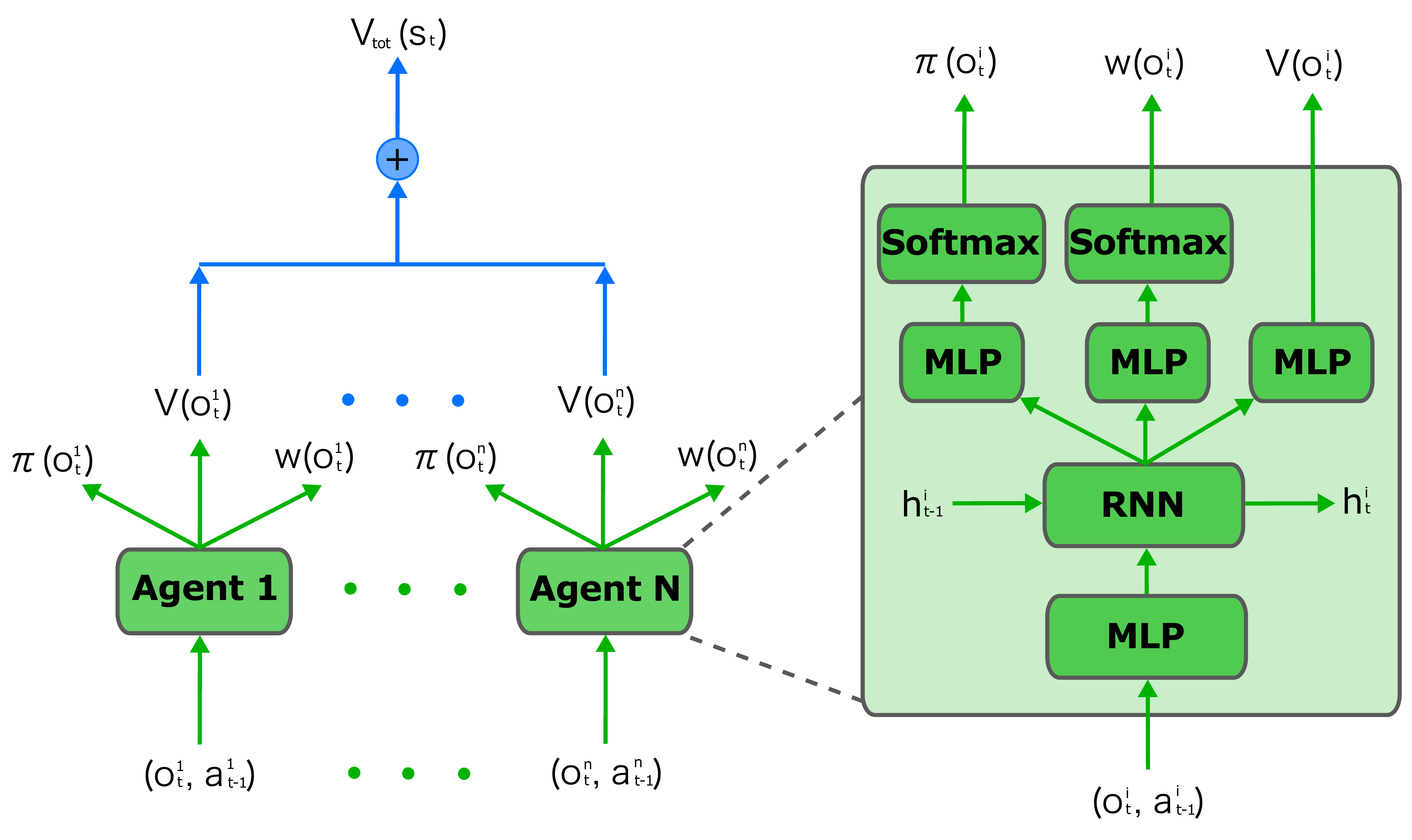}}
\subfigure[IVDAC-mix structure]{
\label{fig:ivadc_mix}
\includegraphics[width=1.2\columnwidth]{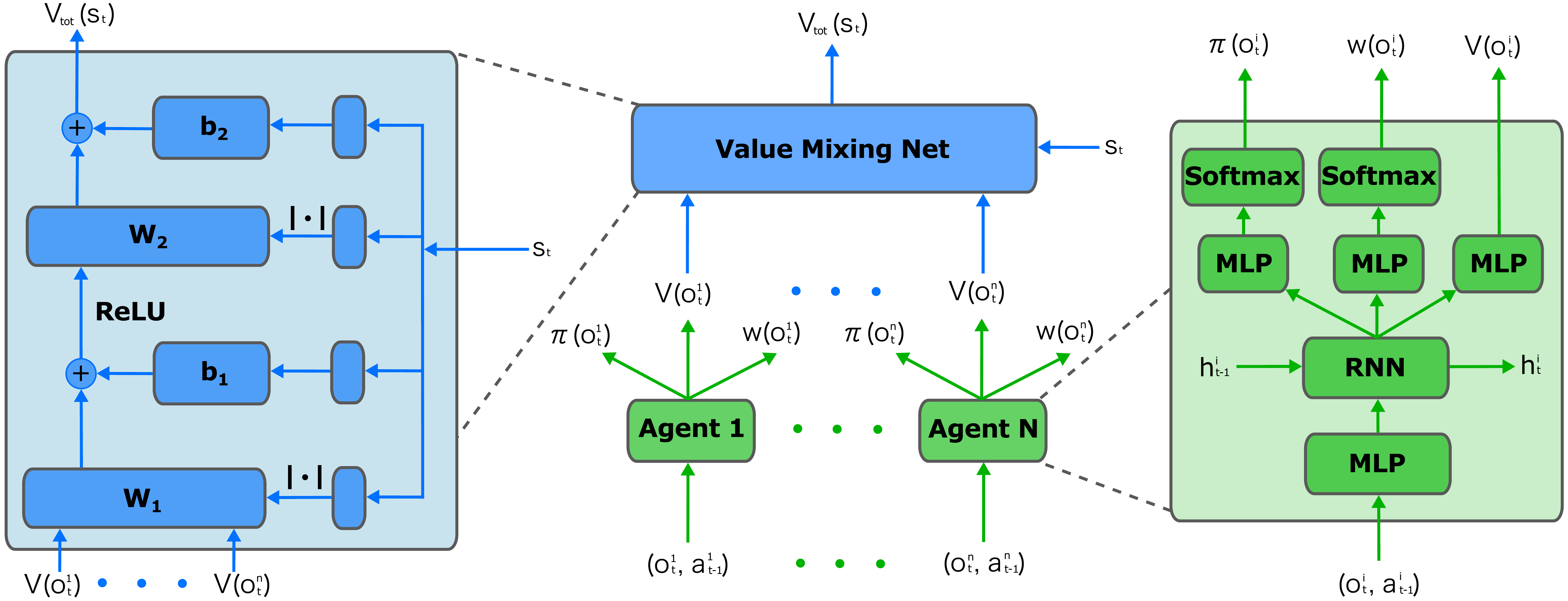}}
\caption{The illustration structure of the IVDAC-sum and IVDAC-mix.}
\end{figure*}

Furthermore, the randomness of the model comes from three sources. The randomness in action is from the policy function: $A \sim \pi(\cdot | s)$; the needs weight function: $W \sim \upsilon(\cdot | s)$ makes the randomness of innate values; the state-transition function: $S' \sim p(\cdot | s, a)$ causes the randomness in state.
In the process, the agent will first generate the needs weight and action based on the current state, then, according to the feedback utilities and the needs weights, calculate the current reward and iterate the process until the end of an episode. Fig. \ref{fig:single_ivrl_single_agent} illustrates the innate-values-driven reinforcement learning (IVRL) model. Fig. \ref{fig:iv_tr} illustrates the trajectory of state $S$, needs weight $W$, action $A$, and reward $R$ in the IVRL model.

\subsection{IVRL Actor-Critic Model}

% Fig. \ref{fig:overview} illustrates the derivation of models from the individual AI agent needs hierarchy based on Maslow's Hierarchy of Needs to innate-values-driven RL (IVRL) \cite{yang2024rationality}.
In the single-agent setting, the IVRL A2C maintains a policy network $\pi(a_t | s_t; \theta)$, a needs network $\omega(w_t | s_t; \delta)$, and a utility value network $u(s_t, a_t; \varphi)$. Since the reward in each step is equal to the current utilities $u(s_t,a_t)$ multiplying the corresponding weight of needs, the state innated-values function can be approximated by presenting it as Eq. \eqref{state_value_net}. The Eq. \eqref{action_grd} and \eqref{needs_grd} show the policy and needs gradient of the Eq. \eqref{state_value_net} deriving $V(s; \theta, \delta)$ according to the {\it Multi-variable Chain Rule}. The policy network $\theta$ and needs network $\delta$ can update the value network $\varphi$ by implementing policy gradient and needs gradient, and using the temporal difference (TD).

\begin{equation}
\begin{split}
    V_{\pi, \omega}(s) & = \sum_{a_t, w_t}\pi(a_t | s_t) \omega(w_t | s_t) \cdot u(s_t, a_t) \\
                       & \approx \sum_{a, w} \pi(a_t | s_t; \theta) \omega(w_t | s_t; \delta) u(s_t, a_t; \varphi) \\
                       & = V(s; \theta, \delta, \varphi)
\label{state_value_net}
\end{split}
\end{equation}
\begin{equation}
\begin{split}
\text{grad~}V(a_t, \theta_t) & = V_\theta(s; \theta, \delta, \varphi) = \frac{\partial V(s; \theta, \delta, \varphi)}{\partial \theta} \\
& = \frac{\partial \pi(a_t | s_t; \theta)}{\partial \theta} \omega(w_t | s_t; \delta) u(s_t, a_t; \varphi)
\label{action_grd}
\end{split}
\end{equation}
\begin{equation}
\begin{split}
\text{grad~}V(w_t, \delta_t) & = V_\delta(s; \theta, \delta, \varphi) = \frac{\partial V(s; \theta, \delta, \varphi)}{\partial \delta} \\
& = \pi(a_t | s_t; \theta) \frac{\partial \omega(w_t | s_t; \delta)}{\partial \delta} u(s_t, a_t; \varphi)
\label{needs_grd}
\end{split}
\end{equation}

% Using an estimate of the utility $u$ function as the baseline function, we subtract the $V$ value term as the advantage value. 
The advantage value equals the difference between the estimate of utility $u$ and the $V$ value.
It presents how much better it is to take a specific action and a needs weight compared to the average general action and the needs weights in the given state Eq. \eqref{adv}. Fig. \ref{fig:single_ivrl_a2c} illustrates the architecture of the Single-agent IVRL A2C model in the StarCraft II.
\begin{equation}
A(s_t, a_t) = U(s_t, a_t) - V(s_t)
\label{adv}
\end{equation}

\subsection{Multi-Agent Policy Gradient Learning}
Multi-agent policy gradient (MAPG) methods extend single-agent policy gradient algorithms with a policy $\pi_{\theta_i}(a_i|o_i)$. However, the main challenges of MAPG are the issues of high variance gradient estimates \cite{lowe2017multi} and credit assignment \cite{foerster2018counterfactual}. Current research usually utilizes centralized training and decentralized execution (CTDE) with a central critic to obtain extra-state information, avoiding high variance in the vanilla MAPG (Eq. \eqref{policy_g}) implementation.
\begin{equation}
\begin{split}
\nabla_{\theta} J = \mathbb{E}_\pi \left[\sum_i \nabla_{\theta} \log \pi_\theta (a_i|o_i) Q_\pi(s, \mathbf{a}) \right]
\label{policy_g}
\end{split}
\end{equation}

One typical method is \cite{lowe2017multi}, which uses a central critic to estimate $Q(s, (a_1, \dots, a_n))$ and optimize actor parameters with a multi-agent DDPG gradient (Eq. \eqref{policy_ddpg}). 
\begin{equation}
\begin{split}
\nabla_{\theta_i} J = \mathbb{E}_\pi \left[\nabla_{\theta_i} \pi (a_i|o_i) \nabla_{a_i} Q_{a_i}(s, \mathbf{a}) | _{a_i = \pi(o_i)} \right]
\label{policy_ddpg}
\end{split}
\end{equation}

On the other hand, \cite{foerster2018counterfactual} demonstrates an approach to solving the credit assignment issue by implementing counterfactual policy gradients (Eq. \eqref{policy_coma}). They claim that the COMA gradients provide agents with tailored gradients that achieve credit assignment with variance reduction.
\begin{eqnarray}
\nabla_{\theta} J = \mathbb{E}_\pi \left[\sum_i \nabla_{\theta} \log \pi (a_i|\tau_i) A_i(s, \mathbf{a}) \right] \label{policy_coma} \\
A_i(s, \mathbf{a}) = Q_\pi(s, \mathbf{a}) - \sum_{a_i} \pi_\theta (a_i|\tau_i) Q_\pi^i(s, (\mathbf{a}_{-i}, a_i))
\end{eqnarray}

\section{Approach Overview}
In the cooperative MAS setting, individual choices of actions and strategies should be mutually consistent in order to achieve their common goals. We extend the single-agent IVRL model to the MAS domain as below.
% Based on the IVRL actor-critic model, we extend it to the MAS RL domain as below.
% Please check background and preliminaries in the Appendix \ref{MAS_IVRL} for more details.

% \subsection{Multi-Agent IVRL Actor-Critic Model}

We propose the innate-value-decomposition actor-critic (IVDAC) by combining the IVRL actor-critic model with the value-decomposition actor-critic (VDAC) \cite{su2021value} method. 
% We extend the single-agent IVRL actor-critic model to work with value-decomposition actor-critic (VDAC) \cite{su2021value} termed innate-value-decomposition actor-critic (IVDAC). 
Here, according to the definition of the {\it difference rewards} \cite{wolpert2001optimal}, we can write the {\it innate-values difference rewards} as Eq. \eqref{difference_r}.
Since it enables agents to learn from a shaped reward (Eq. \eqref{difference_r}), any action taken by an agent $i$ that improves $D_i$ also improves the global reward $r(s, \mathbf{a}, \mathbf{w})$. In other words, the global reward is monotonically increasing with $D_i$.
Where $c_i$ is a default action and $\mathbf{w}$ presents needs weights.
\begin{equation}
D_i = r(s, \mathbf{a}, \mathbf{w}) - r(s, (\mathbf{a}_{-i}, c_i), \mathbf{w})
\label{difference_r}
\end{equation}

Furthermore, we decompose the state innate-value $V_{tot}$ into local-state $V_i$ following relationships as Eq. \eqref{local_values}.
\begin{equation}
\frac{\partial V_{tot}}{\partial V_i}  \geq 0, ~~~ \forall ~ i \in N.
\label{local_values}
\end{equation}

More concretely, when the other agents take action $a_{-i}$ in the local states $o_i$, the agent $i$ chooses any actions $a_i$ to a high innate value in the same local state that will improve the global state innate-value $V_{tot}$. 
Based on the VDAC \cite{su2021value}, we also consider two variants of innate-value-decomposition -- {\it IVDAC-sum} and {\it IVDAC-mix} -- integrating the single-agent IVRL actor-critic model to study their performance. 
% Please check the Appendix \ref{MAS_IVRL} for more details. 

{\it a) IVDAC-sum: } In the IVDAC-sum, accroding to Eq. \eqref{state_value_net}, the global state's innate value $V_{tot}(s)$ can be presented as the summation of the local state's innate values (Eq. \eqref{tot_state_value_net}). The Fig. \ref{fig:ivadc_sum} illustrates the structure. Moreover, the minibatch gradient descent optimizes distributed critic by minimizing the loss Eq. \eqref{ivdac_sum_loss}. Where $s_k$ is the last state and $T$ is the upper bound of $k$.
\begin{equation}
\begin{split}
    V_{tot}(s_t) & = \sum_i V_t^i(o_t^i) \\
    & = \sum_i \sum_{a_t^i, w_t^i}\pi(a_t^i | s_t) \omega(w_t^i | s_t) \cdot u(s_t, a_t^i)
\label{tot_state_value_net}
\end{split}
\end{equation}
\begin{equation}
\begin{split}
    L_t(\varphi) = \bigg( y_t - V_{tot}(s_t) \bigg)^2 = \bigg( y_t - \sum_i V_{\varphi_i} (o_t^i) \bigg)^2
\label{ivdac_sum_loss}
\end{split}
\end{equation}
\begin{equation}
\begin{split}
    y_t = \sum_j^{k-t-1} \gamma^j r_j + \gamma^{(k-t)} V_{tot}(s_k)
\label{y_t}
\end{split}
\end{equation}

Moreover, the policy network is following the policy gradient Eq. \eqref{tot_state_value_net}, \eqref{action_grd}, \eqref{needs_grd}, and \eqref{adv} to update its parameters.

{\it b) IVDAC-mix: }

Like the QMIX \cite{rashid2020monotonic}, the mixing network is a feed-forward neural network that inputs agent network outputs and mixes them monotonically to generate the $Q_{tot}$ values, as shown in Fig. \ref{fig:ivadc_mix}. Specifically, to enforce the monotonicity constraint of the Eq. \eqref{local_values}, the weights (but not the biases) of the mixing network are restricted to being non-negative. This allows the network to approximate any monotonic function arbitrarily well \cite{dugas2009incorporating}. Separate hypernetworks \cite{ha2016hypernetworks} produce the weights of the mixing network, and each hypernetwork takes the state $s$ as input and generates the weights of one layer of the mixing network. Moreover, the hypernetwork uses an absolute activation function to ensure the non-negative output weights and biases are not restricted to being non-negative. Then, a 2-layer hypernetwork with a ReLU activation function following the first layer produces the final bias. Finally, the hypernetwork outputs are reshaped into a matrix of appropriate size.

Here, the mixing network structure (including hypernetworks) also follows the central critic and takes local state innate values $V^a(o^a)$ as additional inputs besides the global state $s$. We can optimize the distributed critics by minimizing the Eq. \eqref{ivdac_mix_loss}. Where $f_{mix}$ presents the mixing network.
\begin{equation}
\begin{split}
    L_t(\varphi) & = \bigg( y_t - V_{tot}(s_t) \bigg)^2 \\
    & = \bigg( y_t - f_{mix} \Big(V_{\varphi_1} (o_t^1), \cdots, V_{\varphi_n} (o_t^n) \Big) \bigg)^2
\label{ivdac_mix_loss}
\end{split}
\end{equation}

Supposing $\psi$ denotes the parameters in the hypernetworks, the corresponding loss function can also be described as $L_t(\psi) = \Big( y_t - V_{tot}(s_t) \Big)^2$, and the update of the policy network follows the same policy gradient in Eq. \eqref{tot_state_value_net}, \eqref{action_grd}, \eqref{needs_grd}, and \eqref{adv}. 

\section{Evaluation through Simulation Studies}
\begin{figure*}[t]
\centering
\subfigure[]{
\label{fig:w_so}
\includegraphics[width=0.65\columnwidth]{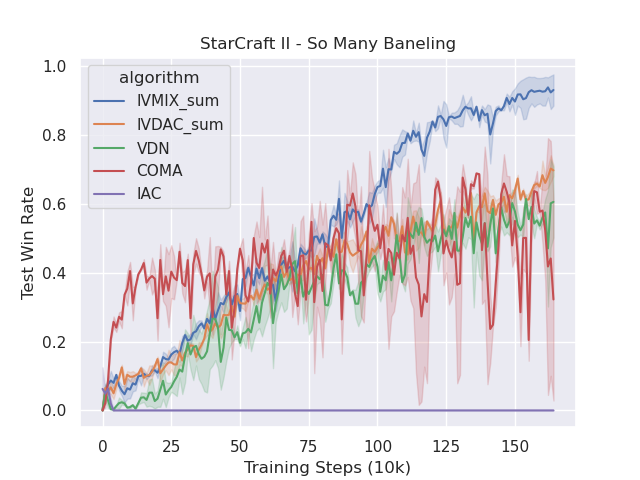}}
\subfigure[]{
\label{fig:w_3m}
\includegraphics[width=0.65\columnwidth]{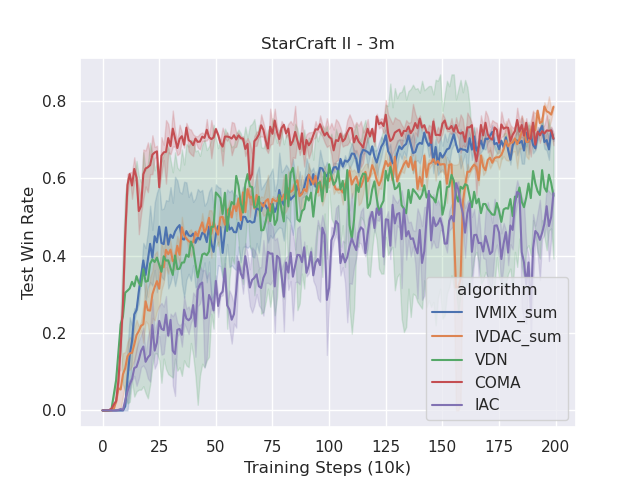}}
\subfigure[]{
\label{fig:w_2s3z}
\includegraphics[width=0.65\columnwidth]{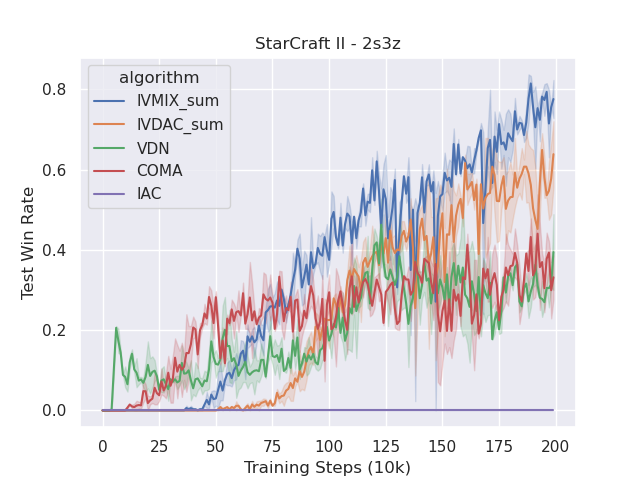}}
\subfigure[]{
\label{fig:d_so}
\includegraphics[width=0.65\columnwidth]{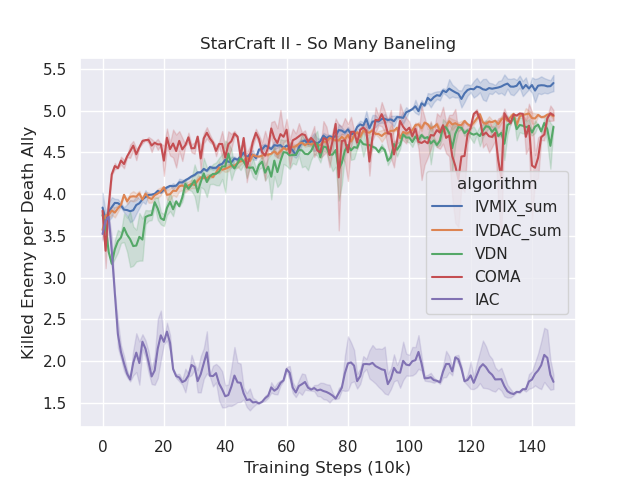}}
\subfigure[]{
\label{fig:d_3m}
\includegraphics[width=0.65\columnwidth]{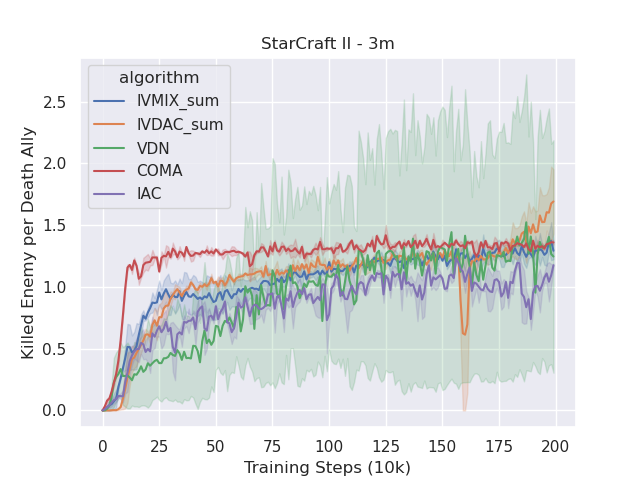}}
\subfigure[]{
\label{fig:d_2s3z}
\includegraphics[width=0.65\columnwidth]{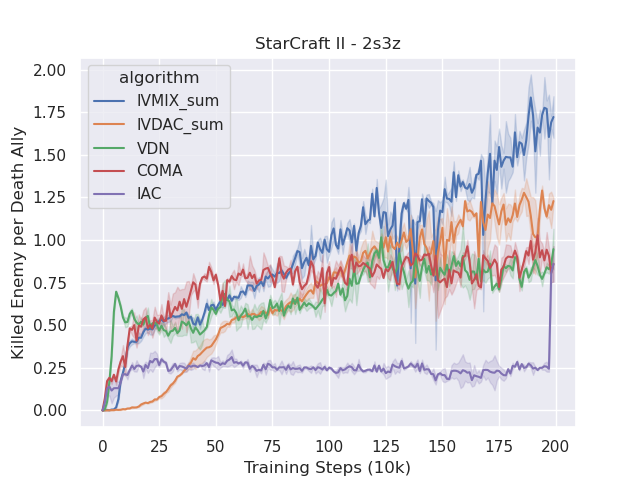}}
\caption{The performance of the dynamic innate-value model based on the IVDAC-sum and IVDAC-mix.}
\end{figure*}

We evaluate the performance of the proposed innate-value-driven multi-agent reinforcement learning architecture IVRL and corresponding models -- IVDAC-sum and IVDAC-mix -- in the StarCraft Multi-Agent Challenge (SMAC) \cite{samvelyan19smac} environment.
% These models were trained on an NVIDIA GeForce RTX 3080Ti GPU with 16 GiB of RAM.
We evaluate the multi-agent IVRL in different categories of combat scenarios (3m, 2s3z, and so many paneling) and analyze their performance as below. 
% Please check experiment setting in the Appendix \ref{MAS_IVRL} for more details.

\textit{a. Homogeneous Close-Quarters Combat (CQC)}

We first study the proposed IVRL architecture with the IVDAC-sum and IVDAC-mix algorithms in the ``So Many Baneling" map, which is a Homogeneous Close-Quarters Combat. The combat is seven Zealots against 32 Baneling cooperatively. The enemy group (Baneling) attacks opponents in a fixed shape (rectangle), which means they cannot change their attacking form. In contrast, the Zealots can transfer their position based on various situations.

Fig. \ref{fig:w_so} shows that the IVDAC-mix has the highest battle-won mean compared with others. Moreover, the IVDAC-sum also performs better than the COMA, VDN, and IAL. It demonstrates that IVDAC-mix and IVDAC-sum can achieve higher and more stable death enemy amounts per unit. Furthermore, Fig. \ref{fig:d_so} also proves that through IVRL, the IVDAC-mix and IVDAC-sum agents can achieve higher enemy death amounts per unit in the homogeneous close-Quarters combat.

\textit{b. Homogeneous Long Range Combat (LRC)}

In the Homogeneous Long Range Combat scenario, we consider two groups of Marines (3 vs. 3) against each other through the guns. The proposed IVRL MAS models also perform better than other methods (Fig. \ref{fig:w_3m} and \ref{fig:d_3m}). However, compared with the homogeneous CQC, it does not present a distinguishing difference from other algorithms. The main reason might be that the utility function design is based on different character categories, game scenarios, etc, which we will study further in future works.

\textit{c. Heterogeneous Melee}

Lastly, we consider the melee in two heterogeneous groups (2 Stalkers and 3 Zealots) with different capabilities. Fig. \ref{fig:w_2s3z} and \ref{fig:d_2s3z} show that the IVDAC-mix and IVDAC-sum can achieve higher winning rates and enemy death amounts per unit. In our experiments, we found that the utility mechanism design plays a crucial and sensitive role in the training process according to various scenarios, which directly decides the performance and sample efficiency. For future work, we will study the utility mechanism representing more organized motivations in the cooperative MAS.

\section{Conclusion and Future Works}

This paper introduces the MAS IVRL model to mimick the complex behaviors in multi-agent interactions. We formulate the model within cooperative multi-agent settings and demonstrate the proposed architecture with the VDAC method in the SAMC environment.
% The results prove that rationally organizing various individual needs can achieve better performance with lower costs in MAS cooperation effectively.
Generally, the innate-value system serves as a unique reward mechanism driving agents to develop diverse actions or strategies satisfying their various needs in the systems. 
% It also builds different personalities and characteristics of agents in their interaction. 
Moreover, organizing agents with similar interests and innate values in the mission can optimize the group utilities and reduce costs effectively, just like ``Birds of a feather flock together." in human society.
For future work, we want to further improve the multi-agent IVRL model and develop a more comprehensive innate-values system to achieve various tasks and test in real multi-robot systems. 

\section{Acknowledgments}
% This work is supported in part by the NSF Foundational Research in Robotics (FRR) Award 2348013 and the Bradley Caterpillar Fellowship 2511133.
This work is supported by the NSF Foundational Research in Robotics (FRR) Award 2348013.

% \clearpage

\bibliography{references}
\bibliographystyle{IEEEtran}

\end{document}